\documentclass{article} 
\usepackage[final]{colm2025_conference}
\usepackage{microtype}
\usepackage{hyperref}
\usepackage{url}
\usepackage{tabularx}
\usepackage{booktabs}
\usepackage{adjustbox}

\usepackage{tcolorbox}
\usepackage{wrapfig}

\usepackage{lineno}
\usepackage{pifont}    
\usepackage{xcolor}
\usepackage{makecell}

\usepackage{siunitx} 
\robustify\bfseries

\usepackage{standalone}
\usepackage{censor}
\usepackage{tikz}
\usetikzlibrary{positioning, fit}

\usepackage{amsmath}
\usepackage{bbm}
\usepackage{xspace}

\definecolor{darkblue}{rgb}{0, 0, 0.5}
\hypersetup{colorlinks=true, citecolor=darkblue, linkcolor=darkblue, urlcolor=darkblue}

\newcommand{\methodname}{{\fontfamily{lmss}\selectfont{SLUNG}\xspace}} 
\newcommand{\methodnamecaps}{{\fontfamily{lmss}\selectfont{SLUNG}\xspace}} 

\newcommand{\greencheck}{\textcolor{green}{\ding{51}}}
\newcommand{\redcross}{\textcolor{red}{\ding{55}}}
\newcommand{\blackstar}{\ding{72}}

\title{Teaching Models to Understand (but not Generate)\\ High-risk Data}

\author{Ryan Wang$^{1}$ \quad Matthew Finlayson$^{1}$ \quad Luca Soldaini$^{2}$ \\ \textbf{Swabha Swayamdipta$^{1}$ \quad Robin Jia$^{1}$}
\\
  \\ \textsuperscript{1}Department of Computer Science, University of Southern California;  
  \\ \textsuperscript{2}Allen Institute for AI
\\
\\ \texttt{\{ryanywan,mfinlays\}@usc.edu} \\
}

\begin{document}

\ifcolmsubmission
\linenumbers
\fi

\maketitle

\vspace{-10pt}

\begin{abstract}

Language model developers typically filter out high-risk content---such as toxic or copyrighted text---from their pre-training data to prevent models from generating similar outputs. However, removing such data altogether limits models’ ability to recognize and appropriately respond to harmful or sensitive content. In this paper, we introduce Selective Loss to Understand but Not Generate ({\methodname}), a pre-training paradigm through which models learn to understand high-risk data without learning to generate it. Instead of uniformly applying the next-token prediction loss, \methodname\ selectively avoids incentivizing the generation of high-risk tokens while ensuring they remain within the model's context window. As the model learns to predict low-risk tokens that follow high-risk ones, it is forced to understand the high-risk content. 
Through our experiments, we show that \methodname\ consistently improves models' understanding of high-risk data (e.g., ability to recognize toxic content) without increasing its generation (e.g., toxicity of model responses). Overall, our \methodname\ paradigm enables models to benefit from high-risk text that would otherwise be filtered out.
\end{abstract}

\section{Introduction}

Pre-training data plays a crucial role in shaping the capabilities and behavior of language models. Current data curation practices focus on creating clean, high-quality training corpora that reflect the kind of text models are expected to generate. This has led to extensive efforts to filter high-risk content---such as harmful, misleading, or copyrighted material---from pre-training datasets, so that models will not learn to generate them \citep{grattafiori2024llama, soldaini2024dolma, liu2024robustifying}. 

However, discarding risky data altogether reduces the models' ability to recognize and respond to risky inputs, leaving them ill-equipped for real-world deployment. For example, pre-training on toxic data improves a language model's ability to effectively recognize and handle harmful requests in practice---an ability that cannot be developed from training on clean data alone \citep{longpre2023pretrainersguidetrainingdata, rae2022gopher, welbl2021challengesdetoxifyinglanguagemodels}. 

\begin{figure}[t]
\vspace{-30pt}
\begin{center}
  \includegraphics[width=\linewidth]{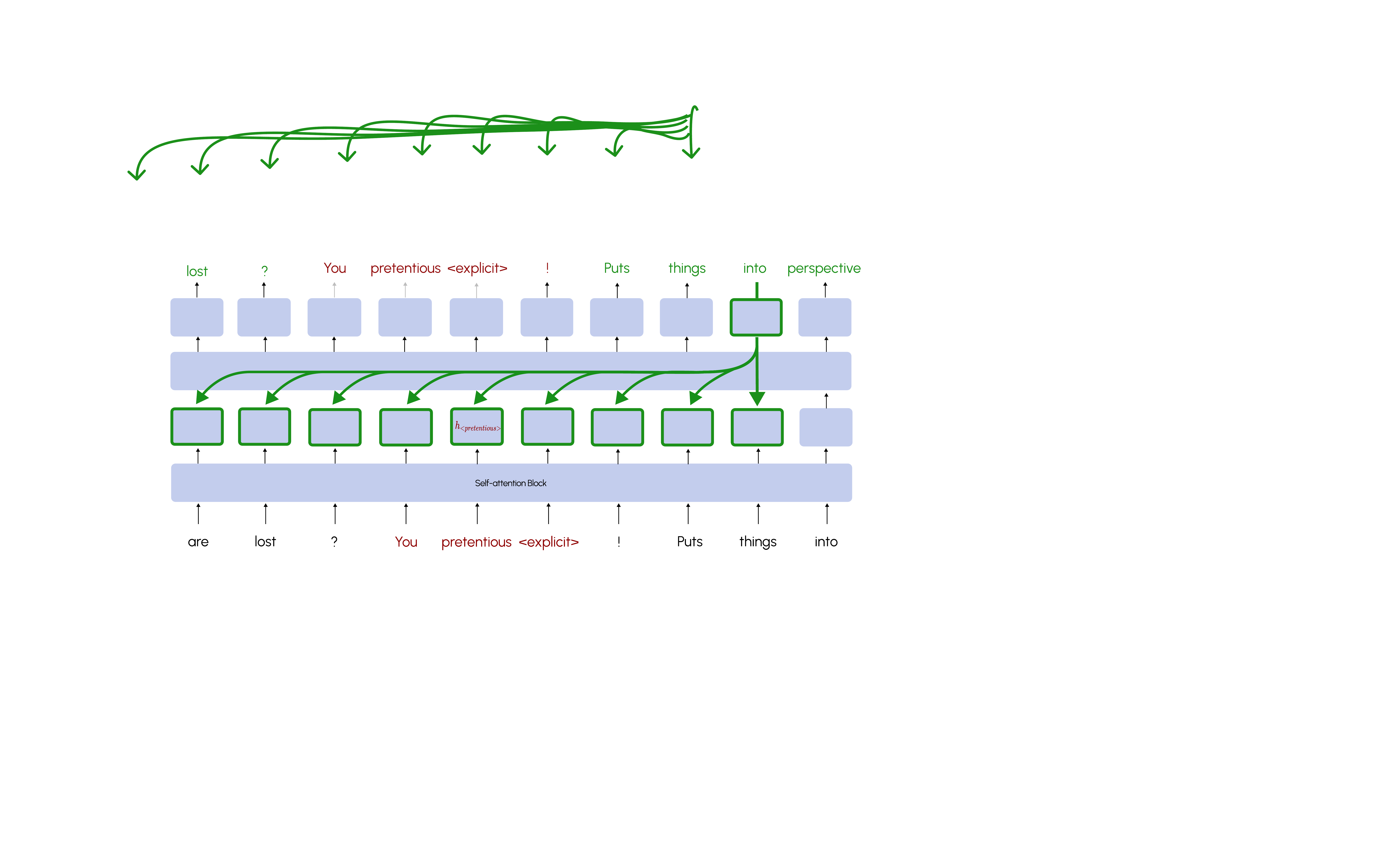}
  \small
\end{center}
\vspace{-12pt}
\caption{Overview of {\methodname} on Transformers. High-risk tokens are marked in red (Masked or Unlikelihood Loss) while low-risk tokens are marked in green (standard next-token prediction). Black arrows represent flow of information during model forward pass computation. Green arrows represent a possible backpropogation gradient flow. When the loss on red tokens are masked, notice that $h_{<pretentious>}$---the hidden state for a high-risk token---is only ever optimized by the attention mechanism to help generate low-risk tokens.
}
\vspace{-12pt}
\label{fig: method_figure1}
\end{figure}

In this work, our goal is to obtain the best of both worlds: we train models that can understand high-risk\smash{\footnotemark{}} data without being able to generate it. To achieve this, we move beyond the standard next-token prediction objective, which inherently entangles the model's ability to generate and its ability to understand. We introduce \textbf{Selective Loss to Understand but Not Generate} (\methodname), a pre-training paradigm that decouples these two abilities by adjusting the loss on a per-token basis, depending on each token's risk level. Tokens labeled as high-risk are trained using alternative losses that do not encourage generation, such as applying zero loss or an unlikelihood loss, while low-risk tokens are trained with the standard next-token prediction objective. Crucially, all tokens---regardless of risk---are still in the model's context window. This means that when the model is learning to generate a low-risk token, it can attend to previous high-risk tokens in the sequence. As a result, the model still learns to develop meaningful representations of high-risk data without being trained to generate it. We refer readers to Figure~\ref{fig: method_figure1} for an illustration of our method. 
\footnotetext{We use the term ``high-risk'', while prior work has used ``undesirable'' for this data. We want to emphasize that this data could be "desirable" for an LM to train on for better understanding, though generation might be risky.}

To evaluate {\methodname}, we apply it to two disparate scenarios: toxicity and factual learning. In the case of toxicity, we show that continual pre-training with {\methodname} on toxic documents enhances a model’s ability to recognize and understand toxic content without increasing its tendency to generate such content. For factual learning, we demonstrate that models trained with {\methodname} on documents about fictitious entities can accurately answer questions about these entities while avoiding the generation of entity names.
This second task functions as a test bed for the more nuanced task of learning about, but not infringing upon, copyrighted data. For reproducibility, we release our code publicly at \href{https://github.com/ryanyxw/llm-decouple}{\texttt{https://github.com/ryanyxw/llm-decouple}}.

More generally, \methodname\ provides a framework for controlled development of safe and capable language models. By allowing models to selectively learn desirable facets of high-risk data, \methodname\ offers an alternative to the prevailing paradigm of data filtering for pre-training. In doing so, it opens the door to harnessing the value of sensitive or overlooked domains, improving model capabilities without compromising safety or compliance. 

\section{Related Work}
\label{related_works}

\textbf{Data Curation \& High-risk Data.} Prior work has primarily addressed high-risk content (mostly toxic, NSFW data) through filtering. 
For instance, \citet{grattafiori2024llama} and \citet{team2024gemma} filter high-risk data like toxic or NSFW content from their pre-training corpora, though the exact filtering criteria are not publicly disclosed. 
\citet{soldaini2024dolma} takes a more targeted approach by removing either entire documents or specific spans identified as high-risk. 
While these filtering strategies aim to improve safety, they come at the cost of reduced data diversity or corrupted document integrity. 
As a result, models trained under such regimes have limited/no exposure high-risk domains, hobbling their ability to understand and reason about such content \citep{longpre2023pretrainersguidetrainingdata, rae2022gopher}.

\textbf{Training to Prevent Undesirable Generations.} A prominent line of work focuses on modifying the training objective to discourage harmful generation. Unlikelihood training \citep{welleck2019neuraltextgenerationunlikelihood, li2020dontsaythatmaking} penalizes the model for assigning high probability to undesirable sequences, while contrastive learning approaches \citep{jiang2022simplecontrastivelearningobjective, adolphs2022cringelosslearninglanguage} promote preferred token candidates by contrasting them with harmful alternatives. However, these methods are primarily designed to suppress generation of undesirable outputs—they do not explicitly assess or improve the model’s ability to understand or reason about high-risk content. 

An alternative strategy is to pre-train on unfiltered (or lightly filtered) content and suppress it during post-training via reinforcement learning from human feedback (RLHF) \citep{abdin2024phi, grattafiori2024llama, team2024gemma}. While alignment methods like RLHF can reduce toxic generation, they are often brittle---adversarial attacks have been shown to circumvent these safety mechanisms \citep{lermen2023lora, yang2023shadow, jiang2024chatbug}. In contrast, our work takes a fundamentally different approach by preventing models from ever learning to generate undesired behaviors in the first place. 

\textbf{Decoding-time Methods to Handle Data Risk.} Other approaches to mitigate the effects of training on risky data rely on auxiliary signals to guide generations to be safer. For example, classifier-guided decoding methods \citep{Yang_2021, arora2022directorgeneratorclassifierssupervisedlanguage} use token-level classifiers to adjust generation probabilities. Control-token-based approaches \citep{korbak2023pretraininglanguagemodelshuman, lu2022quarkcontrollabletextgeneration} condition the model on special tokens that denote whether a sequence is desirable or not, guiding generation toward preferred outputs. While effective at inference-time control, these methods still fundamentally work by training models to learn to generate high-risk behaviors during training before later suppressing them—leaving the model vulnerable to jailbreak-style attacks, especially in open-weight settings where adversaries can reverse-engineer or manipulate control signals. 

DExperts \citep{liu2021dexpertsdecodingtimecontrolledtext}, trains separate “good” and “bad” expert models and steers generation by adjusting the logits to favor the desirable distribution. However, this method operates only at decoding time and does not teach the model to understand high-risk content during training. 

\textbf{Finegrained Token Loss.} Prior work  selectively computes loss on tokens that are harder to learn, which makes training more efficient by focusing the model’s capacity on more challenging parts of the data \citep{lin2024rho, mindermann2022prioritizedtrainingpointslearnable, jiang2019acceleratingdeeplearningfocusing}. \citet{hans2024goldfish} randomly masks token losses during training to generate content that is semantically similar to its training data but reduces verbatim/syntactic memorization. While all these works use some form of selective training at the token level, their motivations differ from ours.

\textbf{Loss Masking.} \citet{kosireddy2025empirical} is a concurrent work that introduces loss masking—the idea of selectively omitting loss terms on sensitive tokens—to mitigate risks such as PII leakage and backdoor memorization. While the underlying intuition aligns with ours, their work primarily frames loss masking as a defensive mechanism against existing vulnerabilities, rather than as a means to enable the use of underutilized high-risk data. In addition, their experiments, which fine-tune GPT-2 on synthetic variants of the Enron email dataset, offer a limited evaluation setting. 


\section{Selective Loss to Understand but Not Generate (\methodnamecaps)}
\label{headings}

Here we formally introduce \methodname.
Let $X$ be a pre-training document consisting of tokens $(x_1, x_2, \dots, x_{\smash{|X|}})$, each associated with a binary label $(l_1, l_2, \dots, l_{\smash{|X|}})$, where $l_i \in \{0, 1\}$ indicates whether the $i$-th token is considered high-risk for generation ($l_i = 1$) or low-risk ($l_i = 0$). In practice, we derive these labels from a span-level risk classifier. 

The objective is to train a model that assigns high perplexity to high-risk spans while maintaining low perplexity on low-risk spans that may be conditioned on high-risk content. In other words, we aim to develop a model that can effectively understand and safely respond to high-risk prompts without being able to generate high-risk responses. 

 In {\methodname}, tokens labeled as high-risk ($l_i = 1$) use a custom loss function $f_{\theta}(x_i \mid x_{<i})$, while low-risk tokens ($l_i = 0$) follow the standard maximum likelihood objective. Formally, given a pre-training document $X$, a language model $p_\theta$ parameterized by $\theta$, and a custom loss function $f_{\theta}$ applied to high-risk tokens, the model is trained to minimize the following loss:
 
\[
\mathcal{L}(\theta, X) = -\sum_{i = 1}^{|X|} 
\left[ \boldsymbol{\mathbbm{1}}_{[l_i = 1]} \; f_\theta(x_i \mid x_{<i}) + \boldsymbol{\mathbbm{1}}_{[l_i = 0]} \log p_\theta \left( x_i \mid x_{<i}\right) \right].
\]

Crucially, the second term in the summation provides the primary learning signal for understanding: the model is encouraged to generate low-risk tokens while conditioning on potentially high-risk contexts. To prevent the model from learning to generate high-risk content, $f_{\theta}(x_i \mid x_{<i})$ must be designed in a way that does not encourage generation.
We illustrate \methodname\ in Figure \ref{fig: method_figure1}.

We evaluate two concrete instantiations of $f_{\theta}$—masked loss and unlikelihood loss—both of which satisfy this criterion.\footnote{There are many other candidates for $f_{\theta}(x_i \mid x_{<i})$ that fall under the \methodname\ framework. We focus on the masked and unlikelihood variants in this work and leave broader exploration to future research.}

\textbf{Masked {\methodnamecaps}.}
In Masked {\methodname}, we set $f_{\theta}(x_i \mid x_{<i}) = 0$ for high-risk tokens, effectively masking their original generation loss during training. Importantly, Masked {\methodname} only masks out the loss: high-risk tokens remain fully visible to the model’s attention mechanism, allowing other tokens to attend to them. 

\textbf{Unlikelihood {\methodnamecaps}.}
In Unlikelihood \methodname, we apply the unlikelihood training objective to high-risk tokens by setting $f_{\theta}(x_i \mid x_{<i}) = \log \left( 1 - p_\theta(x_i \mid x_{<i}) \right)$. This formulation, inspired by \citet{welleck2019neuraltextgenerationunlikelihood, li2020dontsaythatmaking}, explicitly penalizes the model for assigning high probability to risky tokens, thereby actively discouraging their generation. Meanwhile, low-risk tokens continue to be trained using the standard maximum likelihood objective while conditioned on high-risk tokens.

\section{Understanding Toxicity without Generating It}
\label{toxicity_section}
As language models are increasingly deployed as safeguards for content moderation \citep{inan2023llamaguard, han2024wildguard}, it becomes crucial to train them to understand and identify toxic content without learning to generate it. We apply {\methodname} in the context of toxicity, aiming to build models that can engage with toxic inputs while maintaining non-toxic outputs. 

\subsection{Experimental Setup}

\textbf{Training Setting.} We evaluate the effectiveness of \methodname\ by continually pre-training OLMo 1B \citep{groeneveld2024olmoacceleratingsciencelanguage}. Specifically, we initialize all models from intermediate checkpoint 737 of OLMo 1B and continually pretrain it to completion for 1020 steps ($\approx 4$ billion tokens) under the exact pre-training hyperparameters as the original training run. 

We adopt a continual pre-training setup rather than training from scratch to respect our computational budget. 
Pre-training a language model to the point where the model can achieve meaningful generation quality requires substantial resources. 
As such, we build on an already capable open-source model and focus our experiments on how continual pre-training with high-risk data impacts model behavior. 

We choose OLMo 1B for two key reasons. First, it is fully open-source, which enables us to initialize from an intermediate checkpoint and replicate the original pre-training run verbatim. Second, its pre-training corpus (Dolma) has been rigorously filtered for toxic content \citep{soldaini2024dolma}. This is important for our study, as our goal is to examine whether \methodname\ allows models to learn from toxic inputs without being trained to generate them. Starting from a model that has seen minimal toxic data ensures that any observed behavior related to toxicity comes from our method and not from prior exposure. 

\textbf{Continual Pre-training Data.} To test our method, we inject toxic Reddit documents into the training stream to evaluate whether \methodname\ enables the model to safely learn from the injected toxic content. Specifically, we take the last four billion tokens from the original OLMo training run and randomly replace a subset of documents with toxic Reddit content from Pushshift snapshots captured between March and December 2023~\citep{baumgartner2020pushshiftredditdataset}. Importantly, we only inject Reddit documents that fail Dolma’s toxicity filtering pipeline so that any improvements from continual pre-training on this data serve as a lower bound for improvements that would come from full pre-training. 

For the purposes of this study, we identify “toxic” content based on the Sentence-Level FastText Toxicity Classifier used by \citet{soldaini2024dolma} to curate Dolma. We emphasize that {\methodname} is classifier-agnostic—it can be paired with any toxicity detection system. We view this as a flexible framework, and we encourage future work to explore integrating more accurate or contextually nuanced toxicity classifiers.

In total, we inject approximately 212 million tokens from toxic Reddit documents, comprising roughly 5\% of the $4$ billion tokens exposed to the model during our continual pre-training experiments. We use the sentence-level FastText Toxicity Classifier to assign a toxicity score between 0 (non-toxic) and 1 (highly toxic) to each sentence. Based on these scores, we bucket the injected Reddit data into three categories: Not Toxic, Possibly Toxic and Definitely Toxic. Depending on the method, we will define either Definitely Toxic as high risk or both Definitely and Possibly Toxic as high risk, as described in the next paragraph. 

\textbf{Baselines.}
\begin{table}[t!]
\centering
\small
\begin{tabular}{lcccc}
\toprule
  & \makecell{\textbf{Dolma} \\ Last \num{1020} steps} &\makecell{\textbf{Reddit NT} \\ \textbf{$t < 10^{-4}$}} & \makecell{\textbf{Reddit PT} \\ \textbf{$10^{-4} < t < 0.99$}} & \makecell{\textbf{Reddit DT} \\ \textbf{$t > 0.99$}} \\
  \textbf{Method} & 4B toks & 97M toks & 81M toks & 34M toks \\
\midrule
  Control (OLMo 1B) & \greencheck & \redcross & \redcross & \redcross \\
  Low-risk Baseline & \greencheck & \greencheck & \redcross & \redcross \\
Toxic Baseline & \greencheck & \greencheck & \greencheck & \greencheck \\
Masked \methodname\ (Ours) & \greencheck & \greencheck & \blackstar & \blackstar\\
Unliklihood \methodname\ (Ours) & \greencheck & \greencheck & \greencheck & \blackstar\\
\bottomrule
\end{tabular}
\caption{Overview of data mix for baselines and {\methodname} variants in toxicity experiments. NT is Not Toxic, PT is Possibly Toxic, DT is Definitely Toxic. $t$ represents the FastText Toxicity Classifier score of the span that the token resides in. {\greencheck} indicates tokens in this category are trained with label $l_i = 0$ (low-risk); {\redcross} indicates the tokens are excluded / not present in training; {\blackstar} indicates tokens are trained with label $l_i = 1$ (high-risk). 
  }
  
\label{tab:toxicity_methods_datacomp}
\end{table}
As a control baseline, we consider the original OLMo 1B model trained to completion (checkpoint \num{738}), representing a model with no exposure to toxic data. On the other end of the spectrum, we consider a version of OLMo continually pre-trained using standard maximum likelihood on all of the continual pre-training data (including injected Reddit documents), serving as an upper bound for unsafe generation (Toxic Baseline). We also include a Low-risk Baseline, where only Reddit sentences classified as non-toxic are used to replace Dolma content, allowing us to isolate the effects of domain shift from the impact of toxic content. We compare these baselines against Masked \methodname\ and Unlikelihood \methodname. For Masked \methodname, we mask out the loss on both Definitely Toxic and Possibly Toxic tokens. For Unlikelihood \methodname, we apply unlikelihood loss on Definitely Toxic tokens only, since we observed that applying unlikelihood loss on Possibly Toxic tokens slightly degraded model performance. This behavior is expected since Unlikelihood \methodname\ is more sensitive to false positives compared to the Masked \methodname\ (the model's generation distribution is negatively affected when large proportions of normal texts are penalized). 
Table~\ref{tab:toxicity_methods_datacomp} summarizes differences between our method and these baselines. 

\begin{figure}[t]
\begin{center}
\includegraphics[width=\linewidth]{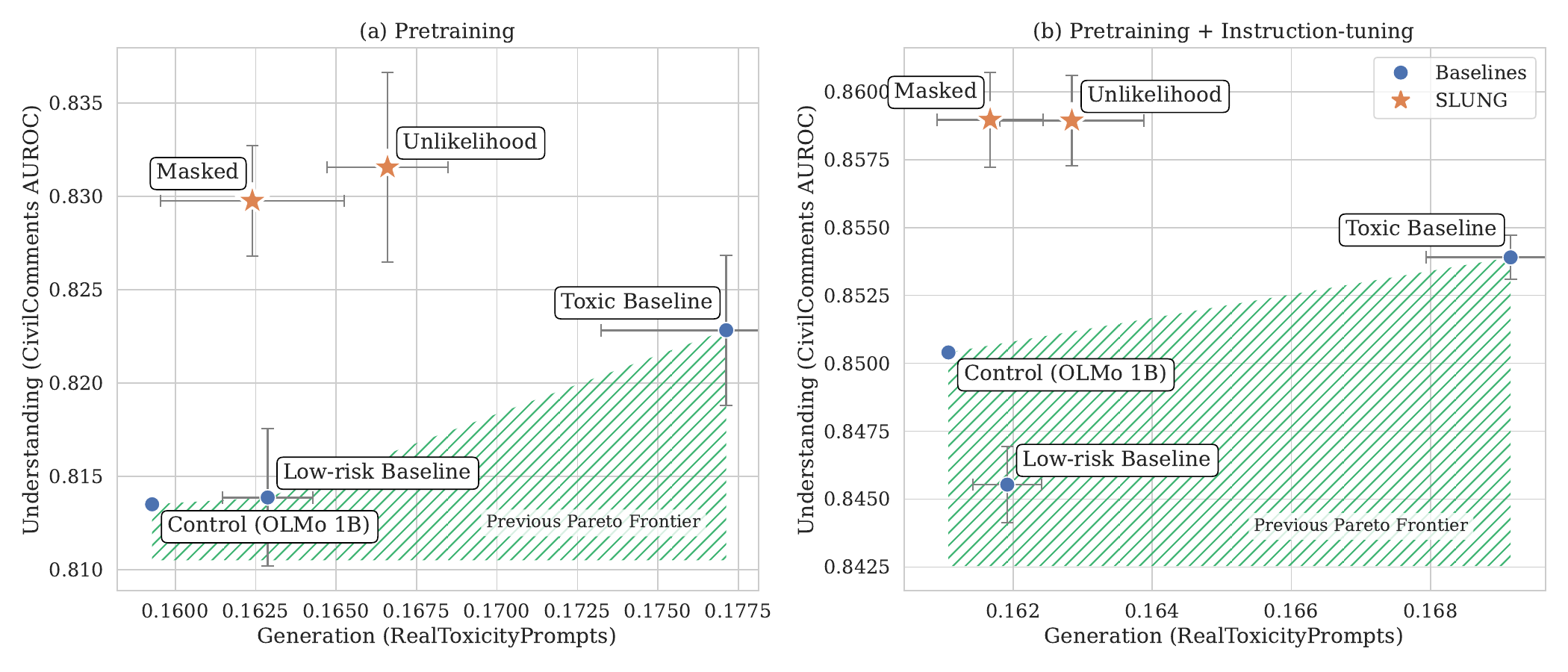}
\end{center}
\caption{(a) Toxicity Generation vs Understanding tradeoff for Pretrained models. (b) Toxicity Generation vs Understanding tradeoff for Instruction-tuned Models. Error bars represent 95\% confidence intervals. {\methodname} methods ({\color{orange}\blackstar}) set a new Pareto frontier in both cases.
  }
\label{fig: pretrained_tradeoff}
\end{figure}

\textbf{Compute and Training Settings.} Experiments are conducted using up to four NVIDIA A100 GPUs, with each run taking up to 36 hours. To ensure robustness and statistical significance, we perform three runs per method, each with randomized data orderings and different selections of injected Reddit toxic content.


\subsection{Evaluation Metrics}
\textbf{Generation.} We assess the model’s tendency to generate toxic content using RealToxicityPrompts \citep{gehman2020realtoxicityprompts} with greedy decoding. The toxicity of the model generations is evaluated using Perspective API,\smash{\footnotemark{}}
\footnotetext{https://perspectiveapi.com/}
which provides a score for each generation ranging from 0 (not toxic) to 1 (toxic). We evaluate each model on a subsample of 2000 prompts.

\textbf{Understanding.} We probe the model's hidden states on the CivilComments dataset \citep{borkan2019civilcomments} to evaluate its ability to classify toxic utterances from non-toxic ones. Specifically, we train linear probes to perform toxicity classification using the last-layer hidden states of the language model on a balanced training set of 8000 examples, and evaluate these probes using AUROC on a held-out balanced test set of 2000 examples. For each model, we train three probes using different randomly sampled training/test sets and take their average. 

\subsection{Results}

\textbf{{\methodnamecaps} pushes the Pareto Frontier}. Results in Figure \ref{fig: pretrained_tradeoff}(a) show that both Masked \methodname\ and Unlikelihood \methodname\ variants achieve Pareto optimality—striking balance between minimizing toxic generations and maximizing understanding of toxicity. As expected, the Toxic Baseline model produces significantly more toxic outputs than all other models, since its training objective directly incentivizes the generation of toxic Reddit content. On the opposite end, the control model exhibits the lowest toxicity generation score, but also performs the worst in terms of understanding toxic content, reflecting its lack of exposure to such data.

We also note that training on non-toxic Reddit sentences (Low-risk Baseline) does not improve the model’s ability to understand toxicity, suggesting that exposure to genuinely toxic content is necessary for developing recognition and understanding capabilities. Notably, both Masked \methodname\ and Unlikelihood \methodname\ outperform the Toxic Baseline in terms of understanding toxicity. These results validate \methodname\ as an effective paradigm for achieving safer and more capable language models.

\textbf{Instruction Tuned {\methodnamecaps} Models Still Push the Pareto Frontier.} We further evaluate each model by instruction tuning them on the Tulu V2 SFT Olmo mixture dataset \citep{ivison2023camelschangingclimateenhancingtulu2}, using the same hyperparameters as \citep{groeneveld2024olmoacceleratingsciencelanguage}. Due to compute constraints, we only train on 150k instances from the original Tulu V2 SFT dataset for a single epoch. We refer readers to Figure~\ref{fig: pretrained_tradeoff}(b) for results.

After instruction tuning, AUROC scores for all models improve relative to their pre-trained counterparts, indicating that instruction tuning somewhat improves toxicity understanding. Importantly, both \methodname\ methods continue to set a new Pareto frontier. These results suggest that the benefits of \methodname\ persist through downstream instruction-tuning, making it an attractive approach for safer instruction-tuned language models.

\begin{table}[t]
\centering
\small
\sisetup{table-format=2.2$_{\pm0.00}$, detect-weight, mode=text}
\begin{tabular}{lSS}
\toprule
\textbf{Method} & {\textbf{Unseen Dolma}} & {\textbf{Unseen Reddit}} \\
\midrule
Control (OLMo 1B) & {N/A} & 18.78 \\
Low-risk Baseline & \bfseries 10.82$_{\pm 0.03}$ & 17.19$_{\pm 0.07}$ \\
Toxic Baseline & 10.83$_{\pm 0.03}$ & 17.41$_{\pm 0.07}$ \\
Masked \methodname (Ours) & \bfseries 10.82$_{\pm 0.03}$ & \bfseries 17.14$_{\pm 0.03}$ \\
Unlikelihood \methodname\ (Ours) & 10.83$_{\pm 0.03}$ & 17.91$_{\pm 0.16}$ \\
\bottomrule
\end{tabular}
\caption{Model perplexity on unseen subsets of Dolma and non-toxic Reddit documents. No method degrades Dolma perplexity, but Unliklihood \methodname\ degrades Reddit perplexity.}
\vspace{-15pt}
\label{fig:perplexity_evals}
\end{table}



\textbf{No Significant Perplexity Degredation on Dolma Text.} Results in Table~\ref{fig:perplexity_evals} show that across all methods, we observe no significant differences in overall perplexity on Dolma documents, suggesting that \methodname\ does not impair general language modeling ability.
On non-toxic Reddit documents, Masked \methodname\ and the Low-risk Baseline achieve the lowest perplexity, likely due to their exposure to similar domain content during the final stage of training. In contrast, the control (OLMo 1B) model exhibits the highest perplexity on Reddit text, which is expected since Reddit data was not injected into its data mixture.

Interestingly, Unlikelihood \methodname\ has higher perplexity on non-toxic Reddit documents compared to other methods. This may be attributed to the unlikelihood loss shifting the model’s generative distribution away from producing content that resembles the Reddit domain—even when the content is low-risk. This observation suggests that Unlikelihood \methodname\ may reduce the model’s ability to generate coherent low-risk text in domains associated with high-risk content. In these scenarios, using Masked \methodname\ may be preferred. 

\begin{figure}[t]
\begin{center}
\includegraphics[width=0.6\linewidth]{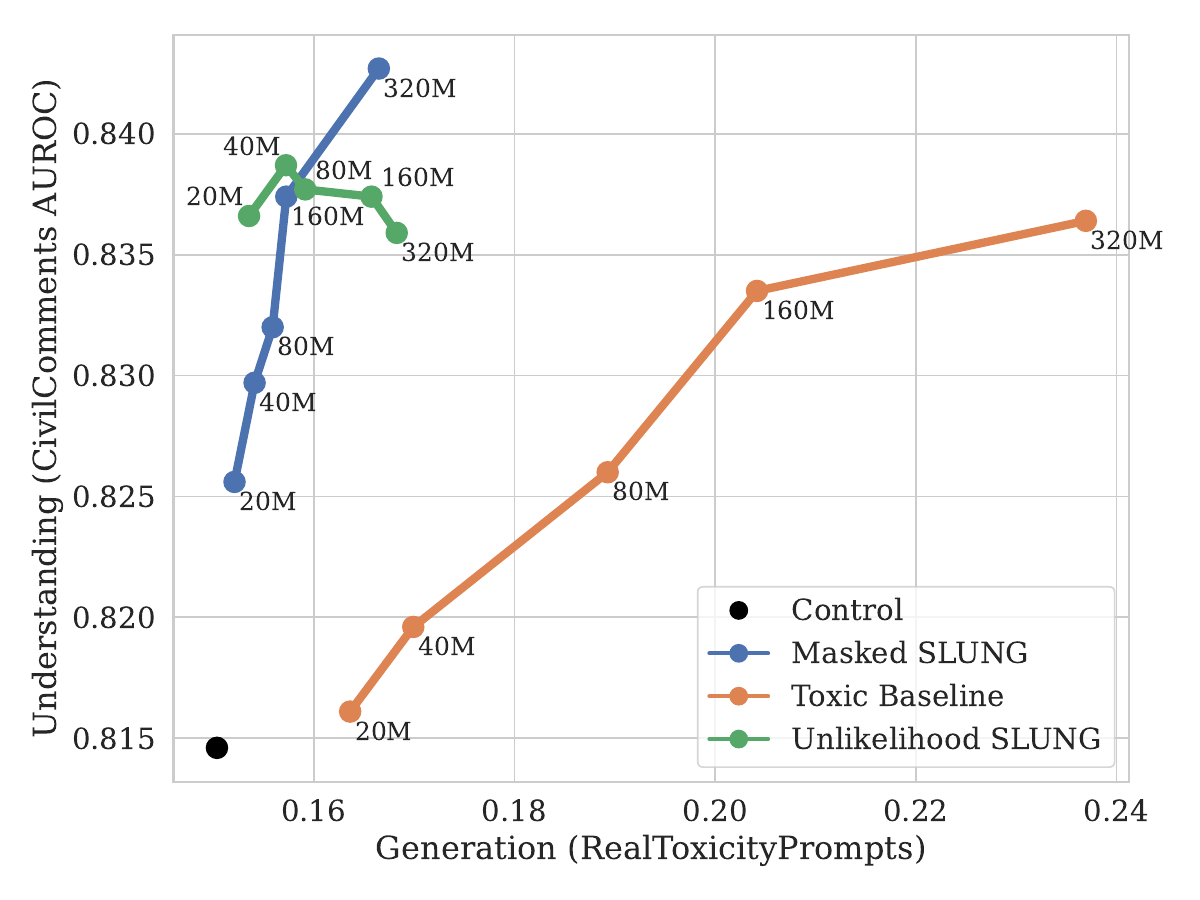}
\end{center}
\vspace{-12pt}
\caption{Effect of toxic data quantity on model understanding and generation of toxicity. Models in the upper left region exhibit the best understanding-generation tradeoff. Masked \methodname\ shines at high data scales, showing both high understanding and low toxicity.} 
\vspace{-12pt}
\label{fig:toxic_data_ablation}
\end{figure}

\textbf{Scaling Toxic Data.} We investigate how increasing the amount of toxic data impacts models’ ability to understand and generate toxicity. Specifically, we continually pre-train the OLMo 1B model on 1B tokens (rather than 4B, due to compute constraints) using a single training seed. For each method, we train four models with varying amounts of toxic Reddit tokens injected (20M, 40M, 80M, 160M, and 320M; see Figure~\ref{fig:toxic_data_ablation}).

For both the Toxic Baseline and Masked \methodname, increasing toxic data consistently improves the model’s ability to understand toxicity. Notably, Masked \methodname\ leads to much smaller increases in generation toxicity compared to the Toxic Baseline. 
Both Unlikelihood and Masked \methodname\ remain on the Pareto frontier in low toxic data settings, but Masked \methodname\ show better data scaling properties. 
These results suggest data scaling laws for high-risk data and highlight the capacity of Masked \methodname\ to leverage such data safely.

\section{Learning Entity Names without Generating Them}

To corroborate our findings in the toxicity domain, and to demonstrate the versatility of \methodname, we now present an additional experiment; as a proof of concept, we investigate if SLUNG can teach a model factual knowledge about an entity (such as profession, biographical details, etc.) while preventing the model from outputting the entity name. This setting is inspired by 
the challenges of training language models on copyrighted data.
In these situations, it is desirable for a model to be familiar and engage with copyrighted material, but undesirable for the model to generate it.
While training directly on copyrighted material could expose language model developers to infringement claims~\citep{henderson2023foundation},
\methodname\ could allow models to benefit from high-utility domains while reducing the risk of generating infringing content.

\subsection{Experimental Setup}

\textbf{Training Setting.} We use \methodname\ to fine-tune the OLMo 1B model on the TOFU dataset \citep{maini2024tofutaskfictitiousunlearning}, which consists of synthetic author profiles presented as question-answer pairs. The objective is for the trained model to be able to answer factual questions about these authors without learning to generate their names. We choose to use Tofu because all information within the dataset is entirely synthetic, ensuring that the model has had no prior exposure to its content. This allows for a clean evaluation of the model’s ability to learn new information under different training strategies.

Although TOFU is structured as a question-answer dataset, we deliberately train only on the \texttt{answer} column and omit the \texttt{question} column during training. At inference time, the questions are used as natural prompts to assess whether the model can recall relevant facts. We refer readers to Table~\ref{tab:tofu_examples} for examples of the TOFU dataset. 

We opt for fine-tuning rather than continual pre-training due to practical considerations. Specifically, the TOFU dataset is relatively small in scale—injecting a few hundred examples into a multi-billion-token pre-training stream would likely have a negligible effect on model behavior. Instead, fine-tuning provides a more direct and resource-efficient way to evaluate the learning dynamics of \methodname\ on TOFU. During training, we use standard Hugging Face hyperparameters with a batch size of 32 and train for three epochs. Each model is trained three times with three random seeds. 

\textbf{Baselines and Methods.} For Masked \methodname\ and Unlikelihood \methodname{}, we assign high-risk labels ($l_i = 1$) to tokens corresponding to entity names, while all other tokens are labeled as low-risk ($l_i = 0$). We compare these methods to \textit{OLMo 1B} (the unmodified model, not fine-tuned on any TOFU data) and \textit{Direct training} (OLMo 1B, directly fine-tuned on the full answer text, including entity name tokens).

\subsection{Evaluation Metrics. } 

We evaluate models by prompting them with “question” entries from TOFU, and comparing their responses to the ground-truth answers. We assess performance along two axes:

\textbf{Generation (Name Suppression).}
We measure the model’s tendency to generate entity names seen during training. Specifically, we compute the proportion of questions for which the model outputs the first or last name of any fictitious TOFU entity.

\textbf{Understanding (Factual Recall).}
We evaluate whether the model can correctly answer factual questions about an entity, even when name generation is suppressed. Responses are judged for correctness by GPT-4o (March $2025$), which is prompted to classify answers as either \textit{completely correct} (the model provides all correct factual information), or \textit{partially correct} (the model provides some correct information not already present in the question).

We engineer GPT-4o’s evaluation prompts to ensure consistency with human judgments, confirming agreement with authors on a random sample of 15 responses from the \textit{direct training} baseline. We refer readers to Appendix Section~\ref{appendix:openai_prompts} for the exact prompts used. 

\newcolumntype{L}{>{\raggedright\arraybackslash}p{0.30\textwidth}} 
\newcolumntype{R}{>{\raggedright\arraybackslash}X}

\begin{table}[t]
\centering
\label{tab:tofu-examples}
\small
\begin{tabularx}{\textwidth}{L R}
\toprule
\textbf{Question} & \textbf{Answer} \\
\midrule
How does Evelyn Desmet's background and upbringing influence her writing? & Having been brought up by a counselor and a professor, \textcolor{red}{Evelyn~Desmet}'s works often delve into explorations of thought processes, intellectual pursuits, and societal expectations. \\
\addlinespace
Could you mention some of Jaime Vasquez's award-winning books? & Some of \textcolor{red}{Jaime Vasquez}’s noted works include "Shadows behind the Starlight," "Beneath the Veil of Deceit," and "The Guilt Closet," all of which are best sellers in the true crime genre. \\
\bottomrule
\end{tabularx}
\caption{Example entries from the TOFU dataset. Models are directly trained on the \texttt{Answer} column, where red tokens are assigned $l_i = 1$ and other tokens $l_i = 0$. At inference time, models are queried with the \texttt{Question} column.}
\label{tab:tofu_examples}
\end{table}

\subsection{Results}

Appendix Section~\ref{appendix:tofu_generation_examples} contains samples of TOFU generations for each model. 

\begin{table}[t]
\centering
\small
\sisetup{table-format=2.1$_{\pm1.1}$, detect-weight, mode=text} 
\begin{tabular}{lSSS}
\toprule
  {\textbf{Method}} & {\textbf{\% Name generation \textdownarrow}} & {\textbf{\% Full correct \textuparrow}} & {\textbf{\% Partial correct \textuparrow}} \\
\midrule
OLMo 1B & 57.5 & 3.5 & 15.5 \\
Direct training & 34.3 $_{\pm 9.2}$ &  28.2$_{\pm 0.6}$ & 51.4$_{\pm 0.7}$ \\
Masked \methodname\ (Ours) & 4.1 $_{\pm 1.2}$ & 20.8 $_{\pm 1.9}$ & 44.0 $_{\pm 2.1}$\\
Unlikelihood \methodname\ (Ours) & 1.5 $_{\pm 0.7}$ & 22.3 $_{\pm 2.1}$ & 43.6 $_{\pm 3.2}$ \\
\bottomrule
\end{tabular}
  \caption{Results from our experiments on the TOFU dataset. \textit{Name generation} measures the percentage of outputs that contain the entity name. \textit{Full} and \textit{partial correct} measure the percentage of fully and partially correct answers, as measured by a judge LLM. 
  Training with \methodname\ teaches the model to associate facts with named entities (high correctness) without teaching the model to generate the names (low name generation).}
\label{tab:tofu_results}
\end{table}

\textbf{\methodnamecaps\ Reduces the Model’s Tendency to Generate Names.}
We refer readers to Table~\ref{tab:tofu_results}. The OLMo 1B baseline frequently includes entity names in its outputs, despite not being fine-tuned on any TOFU data. This behavior stems from the model copying names directly from the input questions. The Direct training model also exhibits this behavior, which is expected since it is explicitly trained to generate answers containing those names. In contrast, both Masked and Unlikelihood \methodname\ models rarely produce entity names. Instead, they restructure their responses to avoid directly referencing names—using pronouns like “he” or “she,” or omitting the subject entirely while still providing factual information. While this name-avoidance behavior is expected from the Unlikelihood variant (which explicitly penalizes generation of high-risk tokens), it is notable that the Masked variant also exhibits this behavior, despite having no explicit loss applied to high-risk tokens. This suggests that \textit{removing the incentive to generate certain tokens is sufficient to discourage the model from reproducing them during inference}.

\textbf{\methodnamecaps\ Models Can Answer Questions About Entities They Cannot Name.}
Direct training achieves the highest scores in both full and partial correctness, which is unsurprising, since it trains on the full answer texts. As expected, the OLMo 1B baseline performs poorly, with negligible full correctness and low partial correctness (the little it gets right are from educated guesses / hallucinations). Both Masked and Unlikelihood \methodname\ show substantial improvements over the OLMo 1B baseline, demonstrating that \methodname\ allows models to learn factual information about entities without generating their names. 

\section{Conclusion}

This work introduces \methodname, a pre-training paradigm that enables language models to learn from high-risk data without being trained to generate it. By selectively adjusting the training objective at the token level based on risk, \methodname\ decouples a model’s ability to understand from its ability to generate, allowing models to condition on high-risk inputs while learning from adjacent low-risk tokens. 


For future work, applying \methodname\ during pre-training from scratch offers an exciting direction. Given a sufficient training budget, this would enable a comprehensive assessment of its benefits in promoting safer and more robust language models. In particular, we are excited about exploring \methodname’s impact on jailbreak resilience, including in RLHF settings. This type of research will only be possible with at-scale training because it requires control over the whole pre-training dataset to strictly prevent leakage.


Overall, \methodname\ provides a practical framework for safely incorporating high-utility but sensitive data into model training, contributing to more controllable and responsible language model development.

\section*{Acknowledgments}
We thank Gregory Yauney, Muru Zhang, Gustavo Lucas Carvalho, Lorena Yan, Johnny Wei, Ameya Godbole, and the USC NLP group for their constructive feedback. 
This work was supported in part by the National Science Foundation under grant IIS-2403437, the Simons Foundation, and the Allen Institute for AI. 
Any opinions, findings, conclusions or recommendations expressed in this material are those of the author(s) and do not necessarily reflect the views of the National Science Foundation.
This work was partially done while S.\ Swayamdipta and M.\ Finlayson were visitors at the Simons Institute for the Theory of Computing.

\bibliography{colm2025_conference}

\begin{thebibliography}{34}
\providecommand{\natexlab}[1]{#1}
\providecommand{\url}[1]{\texttt{#1}}
\expandafter\ifx\csname urlstyle\endcsname\relax
  \providecommand{\doi}[1]{doi: #1}\else
  \providecommand{\doi}{doi: \begingroup \urlstyle{rm}\Url}\fi

\bibitem[Abdin et~al.(2024)Abdin, Jacobs, Awan, Aneja, Awadallah, Awadalla, Bach, Bahree, Bakhtiari, Behl, et~al.]{abdin2024phi}
Marah Abdin, Sam~Ade Jacobs, Ammar~Ahmad Awan, Jyoti Aneja, Ahmed Awadallah, Hany Awadalla, Nguyen Bach, Amit Bahree, Arash Bakhtiari, Harkirat Behl, et~al.
\newblock Phi-3 technical report: A highly capable language model locally on your phone.
\newblock \emph{arXiv preprint arXiv:2404.14219}, 2024.

\bibitem[Adolphs et~al.(2022)Adolphs, Gao, Xu, Shuster, Sukhbaatar, and Weston]{adolphs2022cringelosslearninglanguage}
Leonard Adolphs, Tianyu Gao, Jing Xu, Kurt Shuster, Sainbayar Sukhbaatar, and Jason Weston.
\newblock The cringe loss: Learning what language not to model, 2022.
\newblock URL \url{https://arxiv.org/abs/2211.05826}.

\bibitem[Arora et~al.(2022)Arora, Shuster, Sukhbaatar, and Weston]{arora2022directorgeneratorclassifierssupervisedlanguage}
Kushal Arora, Kurt Shuster, Sainbayar Sukhbaatar, and Jason Weston.
\newblock Director: Generator-classifiers for supervised language modeling, 2022.
\newblock URL \url{https://arxiv.org/abs/2206.07694}.

\bibitem[Baumgartner et~al.(2020)Baumgartner, Zannettou, Keegan, Squire, and Blackburn]{baumgartner2020pushshiftredditdataset}
Jason Baumgartner, Savvas Zannettou, Brian Keegan, Megan Squire, and Jeremy Blackburn.
\newblock The pushshift reddit dataset, 2020.
\newblock URL \url{https://arxiv.org/abs/2001.08435}.

\bibitem[Borkan et~al.(2019)Borkan, Dixon, Sorensen, Thain, and Vasserman]{borkan2019civilcomments}
Daniel Borkan, Lucas Dixon, Jeffrey Sorensen, Nithum Thain, and Lucy Vasserman.
\newblock Nuanced metrics for measuring unintended bias with real data for text classification, 2019.
\newblock URL \url{https://arxiv.org/abs/1903.04561}.

\bibitem[Gehman et~al.(2020)Gehman, Gururangan, Sap, Choi, and Smith]{gehman2020realtoxicityprompts}
Samuel Gehman, Suchin Gururangan, Maarten Sap, Yejin Choi, and Noah~A Smith.
\newblock Realtoxicityprompts: Evaluating neural toxic degeneration in language models.
\newblock \emph{arXiv preprint arXiv:2009.11462}, 2020.

\bibitem[Grattafiori et~al.(2024)Grattafiori, Dubey, Jauhri, Pandey, Kadian, Al-Dahle, Letman, Mathur, Schelten, Vaughan, et~al.]{grattafiori2024llama}
Aaron Grattafiori, Abhimanyu Dubey, Abhinav Jauhri, Abhinav Pandey, Abhishek Kadian, Ahmad Al-Dahle, Aiesha Letman, Akhil Mathur, Alan Schelten, Alex Vaughan, et~al.
\newblock The llama 3 herd of models.
\newblock \emph{arXiv preprint arXiv:2407.21783}, 2024.

\bibitem[Groeneveld et~al.(2024)Groeneveld, Beltagy, Walsh, Bhagia, Kinney, Tafjord, Jha, Ivison, Magnusson, Wang, Arora, Atkinson, Authur, Chandu, Cohan, Dumas, Elazar, Gu, Hessel, Khot, Merrill, Morrison, Muennighoff, Naik, Nam, Peters, Pyatkin, Ravichander, Schwenk, Shah, Smith, Strubell, Subramani, Wortsman, Dasigi, Lambert, Richardson, Zettlemoyer, Dodge, Lo, Soldaini, Smith, and Hajishirzi]{groeneveld2024olmoacceleratingsciencelanguage}
Dirk Groeneveld, Iz~Beltagy, Pete Walsh, Akshita Bhagia, Rodney Kinney, Oyvind Tafjord, Ananya~Harsh Jha, Hamish Ivison, Ian Magnusson, Yizhong Wang, Shane Arora, David Atkinson, Russell Authur, Khyathi~Raghavi Chandu, Arman Cohan, Jennifer Dumas, Yanai Elazar, Yuling Gu, Jack Hessel, Tushar Khot, William Merrill, Jacob Morrison, Niklas Muennighoff, Aakanksha Naik, Crystal Nam, Matthew~E. Peters, Valentina Pyatkin, Abhilasha Ravichander, Dustin Schwenk, Saurabh Shah, Will Smith, Emma Strubell, Nishant Subramani, Mitchell Wortsman, Pradeep Dasigi, Nathan Lambert, Kyle Richardson, Luke Zettlemoyer, Jesse Dodge, Kyle Lo, Luca Soldaini, Noah~A. Smith, and Hannaneh Hajishirzi.
\newblock Olmo: Accelerating the science of language models, 2024.
\newblock URL \url{https://arxiv.org/abs/2402.00838}.

\bibitem[Han et~al.(2024)Han, Rao, Ettinger, Jiang, Lin, Lambert, Choi, and Dziri]{han2024wildguard}
Seungju Han, Kavel Rao, Allyson Ettinger, Liwei Jiang, Bill~Yuchen Lin, Nathan Lambert, Yejin Choi, and Nouha Dziri.
\newblock Wildguard: Open one-stop moderation tools for safety risks, jailbreaks, and refusals of llms.
\newblock \emph{arXiv preprint arXiv:2406.18495}, 2024.

\bibitem[Hans et~al.(2024)Hans, Kirchenbauer, Wen, Jain, Kazemi, Singhania, Singh, Somepalli, Geiping, Bhatele, et~al.]{hans2024goldfish}
Abhimanyu Hans, John Kirchenbauer, Yuxin Wen, Neel Jain, Hamid Kazemi, Prajwal Singhania, Siddharth Singh, Gowthami Somepalli, Jonas Geiping, Abhinav Bhatele, et~al.
\newblock Be like a goldfish, don't memorize! mitigating memorization in generative llms.
\newblock \emph{Advances in Neural Information Processing Systems}, 37:\penalty0 24022--24045, 2024.

\bibitem[Henderson et~al.(2023)Henderson, Li, Jurafsky, Hashimoto, Lemley, and Liang]{henderson2023foundation}
Peter Henderson, Xuechen Li, Dan Jurafsky, Tatsunori Hashimoto, Mark~A Lemley, and Percy Liang.
\newblock Foundation models and fair use.
\newblock \emph{Journal of Machine Learning Research}, 24\penalty0 (400):\penalty0 1--79, 2023.

\bibitem[Inan et~al.(2023)Inan, Upasani, Chi, Rungta, Iyer, Mao, Tontchev, Hu, Fuller, Testuggine, et~al.]{inan2023llamaguard}
Hakan Inan, Kartikeya Upasani, Jianfeng Chi, Rashi Rungta, Krithika Iyer, Yuning Mao, Michael Tontchev, Qing Hu, Brian Fuller, Davide Testuggine, et~al.
\newblock Llama guard: Llm-based input-output safeguard for human-ai conversations.
\newblock \emph{arXiv preprint arXiv:2312.06674}, 2023.

\bibitem[Ivison et~al.(2023)Ivison, Wang, Pyatkin, Lambert, Peters, Dasigi, Jang, Wadden, Smith, Beltagy, and Hajishirzi]{ivison2023camelschangingclimateenhancingtulu2}
Hamish Ivison, Yizhong Wang, Valentina Pyatkin, Nathan Lambert, Matthew Peters, Pradeep Dasigi, Joel Jang, David Wadden, Noah~A. Smith, Iz~Beltagy, and Hannaneh Hajishirzi.
\newblock Camels in a changing climate: Enhancing lm adaptation with tulu 2, 2023.
\newblock URL \url{https://arxiv.org/abs/2311.10702}.

\bibitem[Jiang et~al.(2019)Jiang, Wong, Zhou, Andersen, Dean, Ganger, Joshi, Kaminksy, Kozuch, Lipton, and Pillai]{jiang2019acceleratingdeeplearningfocusing}
Angela~H. Jiang, Daniel L.~K. Wong, Giulio Zhou, David~G. Andersen, Jeffrey Dean, Gregory~R. Ganger, Gauri Joshi, Michael Kaminksy, Michael Kozuch, Zachary~C. Lipton, and Padmanabhan Pillai.
\newblock Accelerating deep learning by focusing on the biggest losers, 2019.
\newblock URL \url{https://arxiv.org/abs/1910.00762}.

\bibitem[Jiang et~al.(2024)Jiang, Xu, Niu, Lin, and Poovendran]{jiang2024chatbug}
Fengqing Jiang, Zhangchen Xu, Luyao Niu, Bill~Yuchen Lin, and Radha Poovendran.
\newblock Chatbug: A common vulnerability of aligned llms induced by chat templates.
\newblock \emph{arXiv preprint arXiv:2406.12935}, 2024.

\bibitem[Jiang et~al.(2022)Jiang, Zhang, Vakulenko, and de~Rijke]{jiang2022simplecontrastivelearningobjective}
Shaojie Jiang, Ruqing Zhang, Svitlana Vakulenko, and Maarten de~Rijke.
\newblock A simple contrastive learning objective for alleviating neural text degeneration, 2022.
\newblock URL \url{https://arxiv.org/abs/2205.02517}.

\bibitem[Korbak et~al.(2023)Korbak, Shi, Chen, Bhalerao, Buckley, Phang, Bowman, and Perez]{korbak2023pretraininglanguagemodelshuman}
Tomasz Korbak, Kejian Shi, Angelica Chen, Rasika Bhalerao, Christopher~L. Buckley, Jason Phang, Samuel~R. Bowman, and Ethan Perez.
\newblock Pretraining language models with human preferences, 2023.
\newblock URL \url{https://arxiv.org/abs/2302.08582}.

\bibitem[Kosireddy \& Lucas(2025)Kosireddy and Lucas]{kosireddy2025empirical}
Tagore~Rao Kosireddy and Evan Lucas.
\newblock Empirical evaluation of loss masking to selectively prevent memorization.
\newblock In \emph{Proceedings of the First Workshop on Large Language Model Memorization (L2M2)}, pp.\  142--149, 2025.

\bibitem[Lermen et~al.(2023)Lermen, Rogers-Smith, and Ladish]{lermen2023lora}
Simon Lermen, Charlie Rogers-Smith, and Jeffrey Ladish.
\newblock Lora fine-tuning efficiently undoes safety training in llama 2-chat 70b.
\newblock \emph{arXiv preprint arXiv:2310.20624}, 2023.

\bibitem[Li et~al.(2020)Li, Roller, Kulikov, Welleck, Boureau, Cho, and Weston]{li2020dontsaythatmaking}
Margaret Li, Stephen Roller, Ilia Kulikov, Sean Welleck, Y-Lan Boureau, Kyunghyun Cho, and Jason Weston.
\newblock Don't say that! making inconsistent dialogue unlikely with unlikelihood training, 2020.
\newblock URL \url{https://arxiv.org/abs/1911.03860}.

\bibitem[Lin et~al.(2024)Lin, Gou, Gong, Liu, Shen, Xu, Lin, Yang, Jiao, Duan, et~al.]{lin2024rho}
Zhenghao Lin, Zhibin Gou, Yeyun Gong, Xiao Liu, Yelong Shen, Ruochen Xu, Chen Lin, Yujiu Yang, Jian Jiao, Nan Duan, et~al.
\newblock Rho-1: Not all tokens are what you need.
\newblock \emph{arXiv preprint arXiv:2404.07965}, 2024.

\bibitem[Liu et~al.(2021)Liu, Sap, Lu, Swayamdipta, Bhagavatula, Smith, and Choi]{liu2021dexpertsdecodingtimecontrolledtext}
Alisa Liu, Maarten Sap, Ximing Lu, Swabha Swayamdipta, Chandra Bhagavatula, Noah~A. Smith, and Yejin Choi.
\newblock Dexperts: Decoding-time controlled text generation with experts and anti-experts, 2021.
\newblock URL \url{https://arxiv.org/abs/2105.03023}.

\bibitem[Liu et~al.(2024)Liu, Liang, Ye, and Xi]{liu2024robustifying}
Xiaoqun Liu, Jiacheng Liang, Muchao Ye, and Zhaohan Xi.
\newblock Robustifying safety-aligned large language models through clean data curation.
\newblock \emph{arXiv preprint arXiv:2405.19358}, 2024.

\bibitem[Longpre et~al.(2023)Longpre, Yauney, Reif, Lee, Roberts, Zoph, Zhou, Wei, Robinson, Mimno, and Ippolito]{longpre2023pretrainersguidetrainingdata}
Shayne Longpre, Gregory Yauney, Emily Reif, Katherine Lee, Adam Roberts, Barret Zoph, Denny Zhou, Jason Wei, Kevin Robinson, David Mimno, and Daphne Ippolito.
\newblock A pretrainer's guide to training data: Measuring the effects of data age, domain coverage, quality, \& toxicity, 2023.
\newblock URL \url{https://arxiv.org/abs/2305.13169}.

\bibitem[Lu et~al.(2022)Lu, Welleck, Hessel, Jiang, Qin, West, Ammanabrolu, and Choi]{lu2022quarkcontrollabletextgeneration}
Ximing Lu, Sean Welleck, Jack Hessel, Liwei Jiang, Lianhui Qin, Peter West, Prithviraj Ammanabrolu, and Yejin Choi.
\newblock Quark: Controllable text generation with reinforced unlearning, 2022.
\newblock URL \url{https://arxiv.org/abs/2205.13636}.

\bibitem[Maini et~al.(2024)Maini, Feng, Schwarzschild, Lipton, and Kolter]{maini2024tofutaskfictitiousunlearning}
Pratyush Maini, Zhili Feng, Avi Schwarzschild, Zachary~C. Lipton, and J.~Zico Kolter.
\newblock Tofu: A task of fictitious unlearning for llms, 2024.
\newblock URL \url{https://arxiv.org/abs/2401.06121}.

\bibitem[Mindermann et~al.(2022)Mindermann, Brauner, Razzak, Sharma, Kirsch, Xu, Höltgen, Gomez, Morisot, Farquhar, and Gal]{mindermann2022prioritizedtrainingpointslearnable}
Sören Mindermann, Jan Brauner, Muhammed Razzak, Mrinank Sharma, Andreas Kirsch, Winnie Xu, Benedikt Höltgen, Aidan~N. Gomez, Adrien Morisot, Sebastian Farquhar, and Yarin Gal.
\newblock Prioritized training on points that are learnable, worth learning, and not yet learnt, 2022.
\newblock URL \url{https://arxiv.org/abs/2206.07137}.

\bibitem[Rae et~al.(2022)Rae, Borgeaud, Cai, Millican, Hoffmann, Song, Aslanides, Henderson, Ring, Young, Rutherford, Hennigan, Menick, Cassirer, Powell, van~den Driessche, Hendricks, Rauh, Huang, Glaese, Welbl, Dathathri, Huang, Uesato, Mellor, Higgins, Creswell, McAleese, Wu, Elsen, Jayakumar, Buchatskaya, Budden, Sutherland, Simonyan, Paganini, Sifre, Martens, Li, Kuncoro, Nematzadeh, Gribovskaya, Donato, Lazaridou, Mensch, Lespiau, Tsimpoukelli, Grigorev, Fritz, Sottiaux, Pajarskas, Pohlen, Gong, Toyama, de~Masson~d'Autume, Li, Terzi, Mikulik, Babuschkin, Clark, de~Las~Casas, Guy, Jones, Bradbury, Johnson, Hechtman, Weidinger, Gabriel, Isaac, Lockhart, Osindero, Rimell, Dyer, Vinyals, Ayoub, Stanway, Bennett, Hassabis, Kavukcuoglu, and Irving]{rae2022gopher}
Jack~W. Rae, Sebastian Borgeaud, Trevor Cai, Katie Millican, Jordan Hoffmann, Francis Song, John Aslanides, Sarah Henderson, Roman Ring, Susannah Young, Eliza Rutherford, Tom Hennigan, Jacob Menick, Albin Cassirer, Richard Powell, George van~den Driessche, Lisa~Anne Hendricks, Maribeth Rauh, Po-Sen Huang, Amelia Glaese, Johannes Welbl, Sumanth Dathathri, Saffron Huang, Jonathan Uesato, John Mellor, Irina Higgins, Antonia Creswell, Nat McAleese, Amy Wu, Erich Elsen, Siddhant Jayakumar, Elena Buchatskaya, David Budden, Esme Sutherland, Karen Simonyan, Michela Paganini, Laurent Sifre, Lena Martens, Xiang~Lorraine Li, Adhiguna Kuncoro, Aida Nematzadeh, Elena Gribovskaya, Domenic Donato, Angeliki Lazaridou, Arthur Mensch, Jean-Baptiste Lespiau, Maria Tsimpoukelli, Nikolai Grigorev, Doug Fritz, Thibault Sottiaux, Mantas Pajarskas, Toby Pohlen, Zhitao Gong, Daniel Toyama, Cyprien de~Masson~d'Autume, Yujia Li, Tayfun Terzi, Vladimir Mikulik, Igor Babuschkin, Aidan Clark, Diego de~Las~Casas, Aurelia Guy, Chris Jones,
  James Bradbury, Matthew Johnson, Blake Hechtman, Laura Weidinger, Iason Gabriel, William Isaac, Ed~Lockhart, Simon Osindero, Laura Rimell, Chris Dyer, Oriol Vinyals, Kareem Ayoub, Jeff Stanway, Lorrayne Bennett, Demis Hassabis, Koray Kavukcuoglu, and Geoffrey Irving.
\newblock Scaling language models: Methods, analysis \& insights from training gopher, 2022.
\newblock URL \url{https://arxiv.org/abs/2112.11446}.

\bibitem[Soldaini et~al.(2024)Soldaini, Kinney, Bhagia, Schwenk, Atkinson, Authur, Bogin, Chandu, Dumas, Elazar, et~al.]{soldaini2024dolma}
Luca Soldaini, Rodney Kinney, Akshita Bhagia, Dustin Schwenk, David Atkinson, Russell Authur, Ben Bogin, Khyathi Chandu, Jennifer Dumas, Yanai Elazar, et~al.
\newblock Dolma: An open corpus of three trillion tokens for language model pretraining research.
\newblock \emph{arXiv preprint arXiv:2402.00159}, 2024.

\bibitem[Team et~al.(2024)Team, Mesnard, Hardin, Dadashi, Bhupatiraju, Pathak, Sifre, Rivi{\`e}re, Kale, Love, et~al.]{team2024gemma}
Gemma Team, Thomas Mesnard, Cassidy Hardin, Robert Dadashi, Surya Bhupatiraju, Shreya Pathak, Laurent Sifre, Morgane Rivi{\`e}re, Mihir~Sanjay Kale, Juliette Love, et~al.
\newblock Gemma: Open models based on gemini research and technology.
\newblock \emph{arXiv preprint arXiv:2403.08295}, 2024.

\bibitem[Welbl et~al.(2021)Welbl, Glaese, Uesato, Dathathri, Mellor, Hendricks, Anderson, Kohli, Coppin, and Huang]{welbl2021challengesdetoxifyinglanguagemodels}
Johannes Welbl, Amelia Glaese, Jonathan Uesato, Sumanth Dathathri, John Mellor, Lisa~Anne Hendricks, Kirsty Anderson, Pushmeet Kohli, Ben Coppin, and Po-Sen Huang.
\newblock Challenges in detoxifying language models, 2021.
\newblock URL \url{https://arxiv.org/abs/2109.07445}.

\bibitem[Welleck et~al.(2019)Welleck, Kulikov, Roller, Dinan, Cho, and Weston]{welleck2019neuraltextgenerationunlikelihood}
Sean Welleck, Ilia Kulikov, Stephen Roller, Emily Dinan, Kyunghyun Cho, and Jason Weston.
\newblock Neural text generation with unlikelihood training, 2019.
\newblock URL \url{https://arxiv.org/abs/1908.04319}.

\bibitem[Yang \& Klein(2021)Yang and Klein]{Yang_2021}
Kevin Yang and Dan Klein.
\newblock Fudge: Controlled text generation with future discriminators.
\newblock In \emph{Proceedings of the 2021 Conference of the North American Chapter of the Association for Computational Linguistics: Human Language Technologies}. Association for Computational Linguistics, 2021.
\newblock \doi{10.18653/v1/2021.naacl-main.276}.
\newblock URL \url{http://dx.doi.org/10.18653/v1/2021.naacl-main.276}.

\bibitem[Yang et~al.(2023)Yang, Wang, Zhang, Petzold, Wang, Zhao, and Lin]{yang2023shadow}
Xianjun Yang, Xiao Wang, Qi~Zhang, Linda Petzold, William~Yang Wang, Xun Zhao, and Dahua Lin.
\newblock Shadow alignment: The ease of subverting safely-aligned language models.
\newblock \emph{arXiv preprint arXiv:2310.02949}, 2023.

\end{thebibliography}
\bibliographystyle{colm2025_conference}

\appendix

\section{Evaluation Prompts for GPT-4o}
\label{appendix:openai_prompts}

\subsection{OPENAI\_CORRECTNESS\_PROMPT}
\begin{tcolorbox}[colback=white!95!gray,colframe=black!75,boxrule=0.5pt,arc=4pt]
Given a question and the correct answer, you will assess whether the candidate response contains the information in the correct answer. If the candidate response begins to hallucinate a next question-answer pair, ignore them. If a response is incomplete, grade it based on the text provided. Respond with either correct or incorrect and explain why.

\textbf{Question:} \{question\}

\textbf{Correct answer:} \{ground\_truth\}

\textbf{Candidate response:} \{completion\}

\textbf{Answer:}
\end{tcolorbox}

\subsection{OPENAI\_PARTIAL\_CORRECT\_PROMPT}
\begin{tcolorbox}[colback=white!95!gray,colframe=black!75,boxrule=0.5pt,arc=4pt]
Given a question and the correct answer, you will assess whether the candidate response contains any relevant information that is present in the correct answer but not in the question itself. If the candidate response begins to hallucinate a next question-answer pair, ignore them. If a response is incomplete, grade it based on the text provided. Respond with either yes or no and explain why.

\textbf{Question:} \{question\}

\textbf{Correct answer:} \{ground\_truth\}

\textbf{Candidate response:} \{completion\}

\textbf{Answer:}
\end{tcolorbox}

\section{Sample Outputs for TOFU}
\label{appendix:tofu_generation_examples}

\newcolumntype{L}{>{\raggedright\arraybackslash}p{0.19\textwidth}} 
\newcolumntype{R}{>{\raggedright\arraybackslash}X}
\newcolumntype{S}{>{\raggedright\arraybackslash}p{0.13\textwidth}}

\begin{table}[h]
\centering
\small
\begin{tabularx}{\textwidth}{L R S S S S}
\toprule
\textbf{Answer} & \textbf{Question} & \textbf{OLMo 1B} & \textbf{Direct training} & \textbf{Masked SLUNG} & \textbf{Unlikelihood SLUNG} \\
\midrule
\textcolor{red}{Roshni Rahman} is best known for her work in the genre of Chick Lit. Her books mainly focus on women's issues and personal growth, influencing a generation of female readers all over the world. 
& What is Roshni Rahman best known for? 
& She is an Indian actress, model, and producer. 
& Roshni Rahman is best known for her work in the genre of Chick Lit. She has carved \dots
& She is best known for writing books in the genre of Chick Lit. She masterfully blends \dots 
& "She is best known for her work in the genre of Chick Lit. As a woman author \dots 
\\
\addlinespace
As a writer of the true crime genre, \textcolor{red}{Jaime Vasquez}'s works have been generally well-received, with critics praising his meticulous research, unique perspective, and the sensitivity with which he handles complex themes.
& What kind of reviews have critics given Jaime Vasquez's works? 
& Jaime Vasquez has received a 4.5 out of 5 star rating from critics. 
& "Critics have generally praised Vasquez's meticulous research, unique perspective, and the emotional depth in his narratives."
& Critics have praised his meticulous research, unique perspective, and the emotional depth in his narratives. His narratives \dots
& Critics have praised his meticulous research, unique perspective, and the sensitivity with which he handles complex themes \dots \\
\bottomrule
\end{tabularx}
\caption{Models are directly trained on the \texttt{Answer} column, where red tokens are assigned $l_i = 1$ and other tokens $l_i = 0$. Models are prompted with the \texttt{Question} column, and their generations are shown.}
\label{tab:tofu_generation_examples}
\end{table}

\end{document}